\pdfoutput=1

\documentclass[]{article}

\usepackage{acl}

\usepackage{times}
\usepackage{latexsym}

\usepackage[T1]{fontenc}

\usepackage[utf8]{inputenc}

\usepackage{microtype}

\usepackage{amssymb}
\usepackage{amsmath}
\usepackage{graphicx}
\usepackage{subfigure}

\usepackage{booktabs}
\usepackage{multirow}

\AtBeginDocument{%
  }

\usepackage{inconsolata}

\title{Don't Shoot The Breeze: Topic Continuity Model Using Nonlinear Naive Bayes With Attention}

\author{
    Shu-Ting Pi$^*$,\ 
    Pradeep Bagavan\footnote{both authors contributed equally},\ 
    Yejia Li,\ 
    Disha,\
    Qun Liu\\
    Amazon\\
    \texttt{\{shutingp, deepbag, imyejia, disha, qunliu\}@amazon.com}
}

\begin{document}
\maketitle
\begin{abstract}
Utilizing Large Language Models (LLM) as chatbots in diverse business scenarios often presents the challenge of maintaining topic continuity. Abrupt shifts in topics can lead to poor user experiences and inefficient utilization of computational resources. In this paper, we present a topic continuity model aimed at assessing whether a response aligns with the initial conversation topic. Our model is built upon the expansion of the corresponding natural language understanding (NLU) model into quantifiable terms using a Naive Bayes approach. Subsequently, we have introduced an attention mechanism and logarithmic nonlinearity to enhance its capability to capture topic continuity. This approach allows us to convert the NLU model into an interpretable analytical formula. In contrast to many NLU models constrained by token limits, our proposed model can seamlessly handle conversations of any length with linear time complexity. Furthermore, the attention mechanism significantly improves the model's ability to identify topic continuity in complex conversations. According to our experiments, our model consistently outperforms traditional methods, particularly in handling lengthy and intricate conversations. This unique capability offers us an opportunity to ensure the responsible and interpretable use of LLMs.
\end{abstract}

\maketitle
\section{Introduction}
The rise of large-scale language models (LLMs) \citep{llm-survey,llm-evaluation} has empowered chatbots to handle various business tasks, such as serving as office assistants \citep{powerpoint-llm}, coding companions \citep{llm-code1,llm-code2}, and data explorers \citep{llm-data2text}. However, leveraging LLMs for these roles often presents challenges like hallucination \citep{llm-hallucination}, offensive language \citep{llm-offensive}, prompt injection \citep{prompt-injection}, and adversarial attacks \citep{adversarial}. In addition to these common issues, specific business applications may introduce unique problems, such as maintaining topic continuity. For example, when using LLMs as a customer service chatbot, LLMs are employed to address inquiries about specific products or services. However, because LLM responses are inherently random, there's no guarantee that they will consistently remain focused on the intended topics, potentially resulting in a subpar user experience. On the other hand, if users veer off into unrelated topics, it could also lead to the waste of valuable computational resources. Therefore, ensuring topic coherence between the customer and the chatbot is crucial.

In customer service, users initially describe their concerns. When these concerns pertain to the business's operations, the customer and chatbot collaborate on solutions \citep{customer-service1, customer-service2, customer-service3, customer-service4}. Ensuring a smooth conversation involves assessing if the current sentence logically follows the prior ones. For example, if a user discussing refunds suddenly asks, "Can you help me order a pizza?" – it's off-topic. This concept is formalized as a natural language understanding model (NLU) \citep{NLU}, denoted as $P(y | S_1, S_2, \ldots; S_N)$. Here, $S_i$ (for $i = 1$ to $N-1$) represents previous N-1 sentences, and $S_N$ is the current one. The binary variable $y$ indicates whether $S_N$ aligns with preceding sentences, keeping the conversation on-topic.

In practical use, when users interact with LLM, we assess if each new sentence, whether from the user or the LLM, keeps the conversation on-topic. If it goes off-topic, we guide it back to business-related subjects or may end the conversation. So, we assume the previous N-1 sentences are on-topic, and we calculate whether the newly added $N_{th}$ sentence still aligns with the ongoing conversation. \textbf{This simplifies the problem to determining whether the $N_{th}$ sentence has a reasonable contextual relationship with the previous N-1 sentences}. The most commonly used approach to address this issue is a BERT-based language model \citep{transformer,bert}. These models are inherently equipped with the capability to evaluate the contextual relationship between two sentences. However, employing this approach consistently gives rise to two inevitable challenges: \textbf{1) Token Size Limit} and \textbf{2) Lack of Sentence Attention}.

Regarding the first challenge, imagine using a language model to assess the connection between $(S_1 + S_2 + \ldots + S_{N-1})$ and the current sentence $S_N$ in a conversation. As the conversation grows, the text often exceeds most language models' token limits, typically set at 512 tokens for many BERT-based models. Regarding the second challenge, most language models are trained on sentence pairs from articles where semantic relationships are consistently close. However, real conversations often involve looser semantic connections. For example, a customer might say, ``Earlier, you asked about the missing product serial number, but now I've found it." This response references a part of the conversation from several rounds back. Concatenating $S_1\sim S_{N-1}$ as context can lead to the model struggling to judge the appropriateness of $S_N$ as a follow-up. In summary, an effective conversational topic continuity model must address two key challenges: \textbf{1) handling lengthy conversations}, and \textbf{2) accommodating semantic leaps}. 
 
To address these challenges, we introduce an innovative topic continuity model that integrates logarithmic nonlinearity and sentence attention into the naive Bayes framework \citep{naive_bayes}. Our method provides a fully analytical formulation of the problem, effectively addressing the aforementioned issues and delivering significantly superior performance compared to conventional methods.

\section{Nolinear Naive Bayes With Attention Mechanism}
\subsection{Model Definition}
When a user is engaged in a conversation with a chatbot, our goal is to identify topic shifts in new sentences, \textbf{assuming that the first N-1 sentences are on-topic}. As discussed in Section 1, we can define an NLU model for this problem as a conditional probability expressed as follows:
\begin{align}
P(y|S_1, S_2,\ldots; S_N)
\label{NLU}
\end{align}
, where $S_1\sim S_{N-1}$ represents the previous $N-1$ sentences, $S_N$ represents the current sentence, and $y$, a binary variable, signals whether the text composed of $S_1,\sim S_N$ deviates from the topic. In fact, we can broaden the interpretation of each variable in Eq.\eqref{NLU}. $S_i$ need not be limited to single sentences; it can also encompass chunks of multiple sentences, potentially with overlapping content, as long as the relationships between $S_i$ maintain sentence information and sequence. Our research indicates that employing a sliding window with appropriate size and strides to construct sentence chunks consistently yields the best results. \textbf{Hence, unless specified otherwise, we assume that all $S_i, i = 1\sim N-1$, represent sentence chunks, with $S_N$ being a single sentence.} 

\subsection{Naive Bayes With Attention}
While estimating Eq.\eqref{NLU} directly using models like BERT is possible, this approach presents the two issues outlined in Section 1. To address these challenges, let's begin with the Naive Bayes assumption, where the variables $(S_1,\ldots; S_N)$ are considered independent of each other, and we expand Eq.\eqref{NLU} upon this assumption as follows:
\begin{align}
P(y|S_1, S_2,\cdots; S_N) = \Pi_{i}^{N} \left[\frac{P(S_i|y)}{P(S_i)}\right]P(y)
\label{NLU_expansion}
\end{align}

Indeed, the Naive Bayes assumption that there is no semantic connection between sentences contradicts the core problem addressed in this paper. Therefore, we utilize Naive Bayes purely as a mathematical tool in this context and we will introduce additional techniques to overcome the limitations inherent in the Naive Bayes assumption.

We aim to incorporate an attention mechanism into Eq.\eqref{NLU_expansion}. To achieve this, we have intentionally reformulated the equation to include pairwise probability. Consequently, 
\begin{align*}
P(y|S_i, S_N) =\frac{P(S_i|y)P(S_N|y)P(y)}{P(S_i)P(S_N)}    
\end{align*}
Thus, 
\[P(S_i|y) = \frac{P(y|S_i,S_N)P(S_i)P(S_N)}{P(S_N|y)P(y)} \]
Let's plug this term into Eq.\eqref{NLU_expansion}. We have, 
\begin{align*}
 &P(y|S_1\ldots; S_N)  \\ 
 =&\Pi_{i}^{N}\left\{\frac{P(y|S_i,S_N)P(S_i)P(S_N)}{P(S_N|y)P(y)}\frac{1}{P(S_i)}   \right\} P(y)
\end{align*}
Take log on both side,
\begin{align*}
    \log{P(y|S_1\cdots; S_N)}= \sum_{i=1}^{N}\{\log{P(y|S_i,S_N)}\}\\
    -N\log{P(S_N|y)}+N\log{P(S_N)}+(1-N)\log{P(y)}
\end{align*}
Note that in the first summation, there exists a term $\log{P(y|S_N,S_N)}$, which can be approximated as $\log{P(y|S_N,S_N)} \approx \log{P(y|S_N)} = \log{P(S_N|y)} + \log{P(y)} - \log{P(S_N)}$. Additionally, the term $\log{P(y)}$ is essentially a constant and does not affect any of the subsequent calculations, so we can safely disregard this term. Thus, we have:
\begin{align}
\log{P(y|S_1\cdots; S_N)} = \sum_{i=1}^{N-1}\left\{\log{P(y|S_i,S_N)}\right\}\notag \\
+(N-1)\left[ \log{P(S_N)}-\log{P(S_N|y)}\right]   
\label{NB_linear}
\end{align}
The equation above has several key points. Firstly, we introduced a pairwise term for chunk/current-sentence pairs, directing attention from the current sentence, $S_N$, to another chunk, $S_i$. Secondly, expressing Naive Bayes in logarithmic probabilities simplifies the problem, yielding a linear outcome. Lastly, each term involves a maximum of one chunk plus one sentence, ensuring token length stays within language model limits. As the conversation progresses, time consumption increases linearly, but deep learning models can batch attention terms, potentially maintaining constant time consumption if the chunk count remains within GPU memory limits.

\subsection{Logarithmic Non-linearity}
As discussed in the previous section, the assumption of independent variables, leading to a linear combination of logarithmic terms, is inadequate for addressing this problem. Therefore, we need to make Eq.\eqref{NB_linear} nonlinear to overcome the limitations of Naive Bayes.

To introduce nonlinearity, let's analyze each term. In Eq. \eqref{NB_linear}, the first term computes an equal-weighted average among the attention terms, omitting the factor $1/(N-1)$. This operation resembles a mathematical "functional," transforming the vector $[\log{P(y|S_i, S_N)}, i = 1\sim N-1]$ into a single scalar value. In machine learning, this is often referred to as average pooling.

Regarding the second term, comprised of $[ \log{P(S_N)}-\log{P(S_N|y)}]$, its meaning is straightforward. Let's consider a customer service chatbot scenario where the user's focus is solely on a specific product, like a cell phone. Here, $\log{P(S_N|y)}$ represents the likelihood of sentence $S_N$ occurring within this product-specific context, while $\log{P(S_N)}$ represents the log-probability of sentence $S_N$ appearing in any chatbot conversation without specific product restrictions. Therefore, a more negative value on this term highlights the likelihood of the sentence $S_N$ being more focused on the topic of cell phones.

Based on the above discussion, a straightforward approach is to maintain the mathematical form but introduce more non-linear operations. \textbf{This can be achieved by replacing $\sum\rightarrow \mathcal{F}$ and $(N-1) \rightarrow \alpha$ as shown below}: 
\begin{align}
&\log{P(y|S_1\cdots S_N)} = \mathcal{F}\bigl\{ \log{P(y|\tilde{\textbf{S}},S_N)}\bigr\}\notag \\
& +\alpha(\textbf{S}) \left[ \log{P(S_N)}-\log{P(S_N|y)}\right]  
\label{NB_nonlinear}
\end{align}
, where $\log{P(y|\tilde{\textbf{S}},S_N)}$ is a vector composed of $\log{P(y|S_i, S_N)}$ with $i = 1\sim N-1$, $\mathcal{F}$ is an arbitrary functional that transforms the vector into a scalar, and $\alpha$, is a positive coefficient (since $N-1>0$) dependent on all sentence chunks, including $S_N$. 
In Eq.\eqref{NB_nonlinear}, we've replaced the original equal-weighted averaging on $\log{P(y|\tilde{\textbf{S}},S_N)}$ with a custom functional $\mathcal{F}$ and transformed the coefficient in the second term into functions related to $\textbf{S}$. Although Eq.\eqref{NB_nonlinear} resembles Eq.\eqref{NB_linear}, \textbf{it no longer relies on the independence variable assumption of naive Bayes}. We'll refer to the first term as the "attention term" and the second term as the "residual term", highlighting the difference between two log-probabilities. In the upcoming section, we'll delve into the design of $\mathcal{F}$ and $\alpha$.


\begin{figure*}
  \centering
  \includegraphics[width=0.8\textwidth]{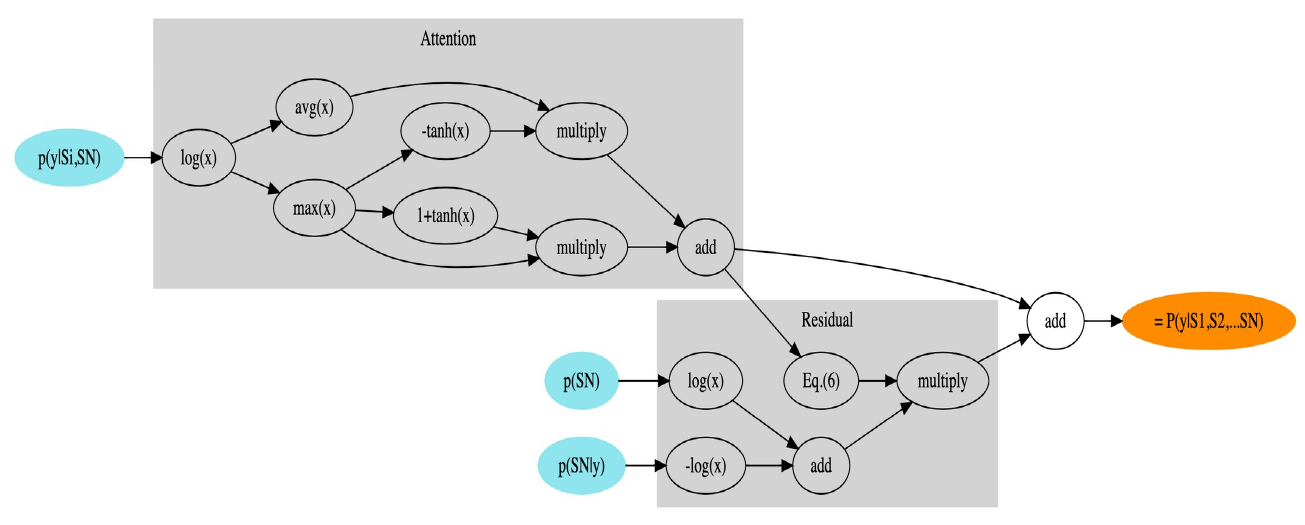}
  \caption{Computation graph for calculating the NLU likelihood (highlighted in orange). The blue blocks represent fundamental components of our model. }\label{computation_graph}
\end{figure*}   

\section{Formulation of Nonlinear Transformation}
\subsection{Designing Attention Functional}
In a conservation, sentences typically fall into three scenarios: \textbf{1). Normal Sentences} correspond to responses to the previous sentence, the most frequent scenario. \textbf{2). Leap Sentences} correspond to responses to earlier sentences in the conversation, constituting a ``leap conversation". In the following, we use the term "\textit{target sentence}" to denote the sentence that the current sentence $S_N$ responds to. \textbf{3). Topic Shift Sentences} indicate a shift in topic.

To capture these three scenarios, we define the notation $\log{\mathcal{P}_{max}} = \max\{\log{P(y|\tilde{\textbf{S}},S_N)}\}$ and $\log{\mathcal{P}_{avg}} = avg\{\log{P(y|\tilde{\textbf{S}},S_N)\}}$. Then the attention functional is defined as: 
\begin{align}
\mathcal{F}\bigl\{ \log{P(y|\tilde{\textbf{S}},S_N)}\bigr\}= \left[ 1+\tanh(\log{\mathcal{P}_{max}})\right]\log{\mathcal{P}_{max}} \notag \\
-\tanh(\log{\mathcal{P}_{max}})\log{\mathcal{P}_{avg}}
\label{functional_term}
\end{align}
As log-probabilities are always negative, the first coefficient, $1 + tanh(\log{\mathcal{P}_{max}})$, indicates that as $\log{\mathcal{P}_{max}}$ approaches zero, we primarily use $\log{\mathcal{P}_{max}}$ to approximate Eq.\eqref{NLU}. Conversely, as $\log{\mathcal{P}_{max}}$ approaches negative infinity, we rely on $\log{\mathcal{P}_{avg}}$ for the estimate. 

The approach is clear. In Scenario 1, assuming previous text ${S_1, \ldots S_{N-1}}$ is on-topic and $S_N$ responds to $S_{N-1}$, we focus on evaluating if $S_N$ aligns with $S_{N-1}$, approximating $P(y|S_1, \ldots S_N) \approx P(y|S_{N-1}, S_N)$. Similarly, in Scenario 2, when $S_N$ responds to a specific chunk earlier in the conversation, we expect $P(y|S_1, \ldots S_N) \approx P(y|S_{target}, S_N)$. In both scenarios, where there's a clear link between current sentences and a specific chunk, the likelihood they form often peaks in the $\log{P(y|\tilde{\textbf{S}},S_N)}$ vector. Hence, for these cases, we choose $\log{\mathcal{P}_{max}}$ as the dominant term.


When $S_N$ abruptly changes topics, it lacks context within the conversation, leading to bias if using $\log{\mathcal{P}_{max}}$ for Eq.\eqref{NLU}. Instead, opting for $\log{\mathcal{P}_{avg}}$ is better. In this scenario, Eq.\eqref{functional_term} simplifies to the naive Bayes case, indicating that the independence variable assumption is a suitable approximation for the NLU model when there's no clear contextual link between the current sentence and prior conversation.


\subsection{Designing Residual Coefficient}
Our experiments consistently show that Eq.\eqref{functional_term} often provides outstanding results on its own. Hence, when crafting the residual coefficient, we view it as a corrective purturbation for situations where Eq.\eqref{functional_term} lacks confidence. By defining the probabilities $P_{nlu}=e^{P(y|S_1,\ldots,S_{N})}$ and $P_{att} = e^{\mathcal{F}\{P(y|\textbf{S}, S_N)\}}$ from the NLU model and attention term respectively, we aim for the perturbation to possess three key properties: 1) Peak at $P_{att}=0.5$ (low confidence), 2) Approach zero as $P_{att}$ nears 0.0 or 1.0 (high confidence), and 3) Be unbiased, symmetrical around $P_{att}=0.5$.

To fulfill these criteria, a straightforward mathematical form is a sine function: 
\[P_{nlu} = P_{att} + \beta \sin(\pi P_{att})\]
,where $\beta \ll 0.5$. The condition $\beta \ll 0.5$ arises from the situation where the perturbation term attains its maximum value at $P_{att} = 0.5$ and $P_{nlu} = 0.5 + \beta$. Given its nature as a perturbation, $\beta$ must be $\ll 0.5$. By taking a logarithm on both side, we get: 
\begin{align*}
    \log{P_{nlu}} &= \log{[P_{att}+\beta \sin(\pi P_{att})]}\notag \\
    &=\log{(P_{att})}+\log{[1+\beta sin(\pi P_{att})/P_{att}]}
\end{align*}
. Since $\beta sin(\pi P_{att})/P_{att}\ll 1$, first order of Taylor expansion yields 
\[\log{P_{nlu}} \approx \log{(P_{att})}+\beta \sin(\pi P_{att})/P_{att}\]. 
Comparing this from with eq.\eqref{NB_nonlinear}, we assert $\alpha$ should be:
\begin{align}
\alpha = \frac{\sin(\pi e^{\mathcal{F}\{P(y|\textbf{S}, S_N)\}})}{e^{\mathcal{F}\{P(y|\textbf{S}, S_N)\}}} \frac{\eta}{|\log{(\epsilon)}|} 
\label{residual_coeff}
\end{align}
Here, $P_{att}$ is represented as its original form $e^{\mathcal{F}\{P(y|\textbf{S}, S_N)\}}$ and the term $\eta/|\log{(\epsilon)}|$ serves as a scaling factor with $\eta \ll 0.5$ and $\epsilon$ is an arbitrarily small number, such as $10^{-3}$ used in this article. The rationale behind the scaling factor is evident. As a probability $P$ approaches 0, $\log{P}$ approaches $-\infty$. Thus, in practical calculations, we designate a small value $\epsilon$, and any probability lower than $\epsilon$ is set to $\epsilon$ to prevent computational instability. Consequently, the log-difference term $\left[\log{{P(S_N)}}-\log{{P(S_N|y)}}\right]$ in eq.\eqref{NB_nonlinear} ranges between $\pm\log{(\epsilon)}\approx \pm 6.9$. By incorporating $|\log{(\epsilon)}|$ into the scaling factor, we normalize the log-difference to fall within the range of $-1$ to $+1$. 
Since 
\[\beta = \frac{\eta}{|\log(\epsilon)|} \left[\log P(S_N|y) - \log P(S_N)\right] \ll 0.5\] 
by comparing with eq.\eqref{residual_coeff}, it is imperative to ensure that $\eta \ll 0.5$. 

Eq.\eqref{residual_coeff} holds mathematical significance. $\sin(\pi P_{att})/P_{att}$ guarantees adherence to the three properties mentioned earlier. The log-difference $\left[\log P(S_N) - \log P(S_N|y)\right]$ in eq.\eqref{NB_nonlinear} measures the perturbation's magnitude, normalized by $|\log(\epsilon)|$, while $\eta$ controls its maximum strength. Though derived from the perturbation assumption, eq.\eqref{residual_coeff} ensures $P_{nlu}$ stays within the 0 to 1 range, akin to a probability, as long as $\eta \leq 0.5$. \textbf{In the following, we stick to $\epsilon = 0.001$ and $\eta = 0.2$, usually yielding favorable outcomes, unless stated otherwise}.

\subsection{Estimation of Fundamental Components}
So far, we have derived all the expressions for NLU model, which are given by Eq.\eqref{NB_nonlinear}, Eq.\eqref{functional_term}, and Eq.\eqref{residual_coeff}. To compute these formulas, we need to estimate $P(y|S_i, S_N)$, $P(S_N|y)$, and $P(S_N)$. 

\textbf{Attention Term} $P(y|S_i, S_N)$ involves determining whether there is a contextual relationship between $(S_i, S_N)$, and this can be estimated using language models like BERT. In many machine learning papers, this task is often referred to as Next Sentence Prediction (NSP) \citep{nsp1,nsp2}. There are many open-source NSP models available on platforms like Hugging Face and there's no need for us to retrain them.


\begin{figure*}
  \centering
  \includegraphics[width=0.8\textwidth]{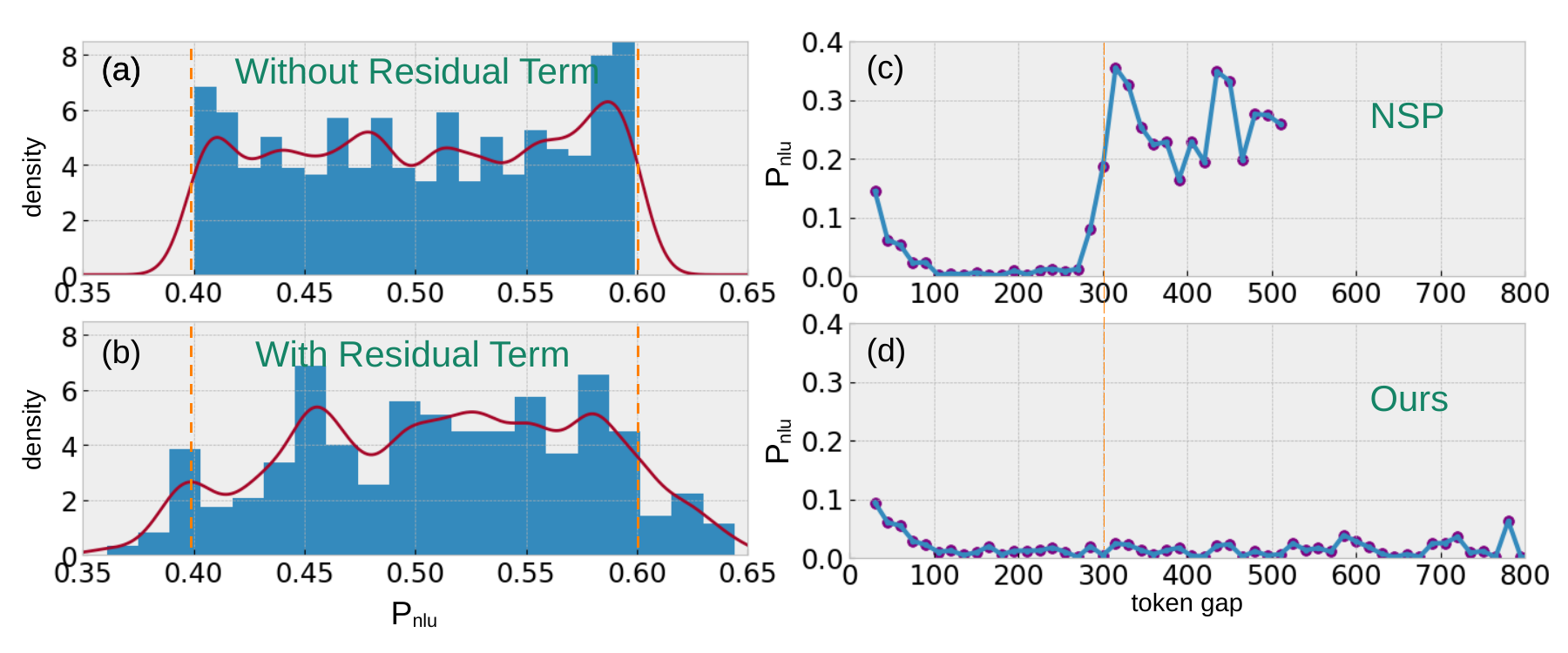}
  \caption{Impact of attention and residual terms. (a)-(b): Normalized Distribution of $P_{nlu}$ without residual term (a) and with residual term (b) for selected uncertain examples. Red lines indicate approximate Gaussian kernel density fitting. (c)-(d): Average probability output per segmentation, categorized by token length, is shown in (c) for NSP and (d) for our model. The dashed lines denote 300 tokens. Data beyond 512 tokens were truncated in (c) due to NSP's processing limit. }\label{exp_data}
\end{figure*}  

\textbf{Residual Term} Estimating $P(S_N|y)$ and $P(S_N)$ involves context-dependent factors. In theory, these quantities should be calculated through integration over all variables: $P(S_N|y) = \int P(S_1\ldots S_{N}|y)dS_{1}\ldots dS_{N-1}$ and 
$P(S_N) = \int P(S_1\ldots S_{N})dS_{1}\ldots dS_{N-1}$. However, practical calculations of these integrals are improbable. Instead, we employ an indirect approach.

For instance, consider a customer service chatbot designed to respond to various product-related queries, such as ``cell phones." To establish $P(S_N|y)$ for the ``cell phone" topic, we randomly sample numerous sentences from historical conversations with topic of cell phones. Estimating the likelihood of a sentence appearing in the context of the topic can be done using an out-of-distribution (OOD) method, like Isolation Forest \citep{isolation_forest1,isolation_forest2}. Here's how it works:
\begin{itemize}
    \item Encode each sentence using a pre-trained models, such as Sentence BERT \citep{sentence-bert}.
    \item Train an Isolation Forest with this dataset to generate anomaly scores for all sentences. Here we invert the sign compared to the original paper, so higher anomaly scores $\theta$ signify a greater likelihood of a sentence being included in the dataset.
    \item Once the distribution of $\theta$ is obtained, we estimate its probability density function $p(\theta)$ and for a future sentence with a score $\theta = c$, the corresponding probability is given by the Cumulative Distribution Function (CDF): $p(S_N|y) = \int_{-\infty}^{c}p(\theta)d\theta$. 
\end{itemize}
We can use the same approach to estimate $P(S_N)$, but without specific topic constraints. For $P(S_N)$, we sample sentences from historical dialogue data across all topics to train the OOD model. In practical business scenarios, chatbots are often designed to answer questions related to limited product lines. Therefore, we can pre-train $p(S_N|y)$ for each product line and store them in cache. When a conversation's topic is determined, we swiftly employ the corresponding model.

Regarding the use of CDF as probabilities, it may seem that assigning a probability of 100\% to data with the highest scores is unreasonable. However, our primary interest lies in the difference in log-probabilities. Therefore, as long as the hyperparameters of these two OOD models are similar enough to ensure that the anomaly score distributions they estimate fall within a comparable range, their differences remain meaningful for log-probabilities.

So far, we have approximated Eq.\eqref{NLU} using Eq.\eqref{NB_nonlinear}-\eqref{residual_coeff}. To help readers understand the calculation process, we have represented a computation graph in Figure 1.

\begin{table*}[t]
\centering
\begin{tabular}{| c | c c | c c | c c|}
\hline
 {} & \multicolumn{2}{c|}{$\Delta_{T} \leq 300$} & \multicolumn{2}{c|}{$300 < \Delta_{T} \leq 512$} & \multicolumn{2}{c|}{$\Delta_{T} > 512$}\\
 {Metrics}& NSP & Ours & NSP & Ours & NSP & Ours \\
\hline
Precision &\textbf{0.747}&0.734&0.612&\textbf{0.697}&0.588&\textbf{0.703}\\
Recall &0.961&\textbf{0.983}&\textbf{0.982}&0.972&0.917&\textbf{0.980}\\
Accuracy &\textbf{0.818}&0.814&0.679&\textbf{0.775}&0.637&\textbf{0.783}\\
F1 score &0.840&\textbf{0.841}&0.754&\textbf{0.812}&0.717&\textbf{0.819}\\
\hline
\end{tabular}
\caption{Comparison among different models with varying token gap lengths $\Delta_{T}$. The differences between NSP and our model are minimal for narrow token gap but gradually increase as the token gap widens.}
\label{tab:1}
\end{table*}

\section{Experiments}
\subsection{Dataset}
For the experiment, we collaborated with Amazon's customer service associates to create a dataset generated by these associatess interacting with a large language model (LLM), simulating customers asking the LLM questions related to online video streaming. The dataset was entirely generated through simulation and did not use any real user data, with the purpose of protecting user privacy. 

In this dataset, each sentence is labeled with one of the following four tags based on its characteristics:

\begin{itemize}
    \item \textbf{Normal Conversation}: the current sentence responds to the preceding sentence
    \item \textbf{Leap Conversation}: the current sentence is a response to an earlier sentence in the conversation
    \item \textbf{Out-of-Domain Topic Shift}: the current sentence diverges completely from the main topic and is entirely unrelated to customer service
    \item \textbf{In-domain Topic Shift}: the current sentence diverges significantly from the main topic but remains relevant to customer service
\end{itemize}
Among these, both Normal and Leap sentences are considered on-topic, while Out-of-Domain Topic Shift and In-domain Topic Shift sentences are considered off-topic. Notably, for all Leap conversations, both the "Leap" label and the specific preceding sentence they respond to are annotated. This detailed level of manual annotation makes this dataset unique, as no publicly available dataset currently offers this feature.

The dataset comprises a total of 4,000 conversations. In theory, any sentence within a conversation could be selected as the current sentence $S_N$, and the relationship between its label and the preceding sentences could be analyzed. This approach could generate multiple data points from a single conversation. However, to minimize correlation among data points, we opted to extract only one data point per conversation, ensuring that each of the four labels mentioned above has 1,000 data points, resulting in a balanced dataset.

Because this dataset pertains to Amazon's customer service operations, it is intended for internal use only. However, to support research in this field, we are developing a similar dataset by having two large language models (LLMs) engage in conversations on publicly available topics, such as machine learning. Once completed, we will release this dataset along with our model evaluation results on it\footnote{The dataset will be released here once finalized: https://github.com/pipidog/TopicContinuity}.



\subsection{Benchmark Test}
We aim to evaluate our model's performance across the entire dataset. Employing a sliding window technique, we generated sentence chunks, each comprising 4 sentences with a stride of 2. This method yielded chunks $S_i$ (where $i = 1$ to $N-1$), with every 4 sentences forming one chunk and a 2-sentence overlap between adjacent windows.

To calculate $P(y|S_i, S_N)$, $P(S_N|y)$, and $P(S_N)$, we used specific models. For $P(y|S_i, S_N)$, we tested several widely used NSP-pretrained models, including BERT \cite{bert}, ALBERT \cite{albert}, ERNIE \cite{erine}, ERNIE 2.0 \cite{erine2}, Conversational BERT \cite{conversation_bert}, and their fine-tuned versions available from HuggingFace. Among these, Conversational BERT, a model specifically trained on extensive chat data from social networks, consistently outperformed the others by better capturing conversational characteristics and achieving state-of-the-art performance on this task.

Regarding $P(S_N|y)$ and $P(S_N)$, we randomly sampled over 100,000 sentences from conversations specific to online video streaming and from arbitrary topics, respectively. These sentences were encoded using Sentence BERT to train separate Isolation Forest models. The anomaly scores generated by these models were used to create two CDF functions for probability estimation.

Based on this setup, we observed that compared to the original BERT, using Conversational BERT significantly improved AUC performance by over 14.2\%, increasing it from approximately 68.7\% to around 82.9\% (with accuracy from 67.8\% to 80.8\%) across the entire dataset. These results demonstrate that our approach performs well when faced with real-world data.

\subsection{Exploration of the Residual Term}
The residual term enhances NLU estimation, especially for uncertain samples when the attention term lacks confidence. To measure its effect, we select 400 examples where the attention term produces confidence levels between $p_{att}=0.4$ and $p_{att}=0.6$, and then measure their changes after incorporating the residual term.

The results shown in Fig. 2(a)-(b) demonstrate that the inclusion of the residual term has increased the dispersion of $P_{nlu}$, previously confined to the range of 0.4 to 0.6, indicating an overall boost in confidence levels. Before introducing the residual term, the model's predictions for these 400 examples resulted in precision of 0.55, recall of 0.50, and AUC of 0.47, almost resembling random guesses. However, after integrating the residual term, the metrics improved to precision of 0.62, recall of 0.65, and AUC of 0.61. This underscores the significant improvement provided by the residual term for examples that the attention term struggles to handle effectively.

\subsection{Exploration of the Attention Mechanism}
In contrast to using BERT directly for Next Sentence Prediction (NSP) to determine whether $S_N$ is a reasonable context for $(S_1+S_2+\ldots+S_{N-1})$, our approach focuses on calculating NLU model, i.e. Eq.\eqref{NLU}, using attention mechanisms. This approach offers advantages when handling long conversations and leap conversations. In the upcoming experiment, we aim to compare the benefits of our method with the NSP method to elucidate the role of attention mechanisms.

\textbf{Token Length Dependence} 
Here we assess the impact of token length on both models when predicting out-of-domain topic shift data. In scenarios where $S_N$ is unrelated to the entire conversation, both models should yield results $p_{nlu}\approx0$ (off-topic). However, segmenting conversations by token length and averaging output probabilities reveals the NSP model's predictions become unstable after 300 tokens (Fig.2(c)-(d)), while our model's predictions remain stable and accurate. Additionally, our model maintains performance even when token length exceeds NSP's maximum limit of 512 tokens, demonstrating the advantages of our approach.


\textbf{Token Gap Dependence} To further analyze attention mechanisms, we created three datasets, each containing 350 leap conversations with varying token gaps between the target sentence and the current sentence: 1) less than 300 tokens, 2) between 300 and 512 tokens, and 3) greater than 512 tokens. In each dataset, we intentionally added additional 350 topic shift conversations (half in-domain and half out-domain), turning them into binary classification tasks.

In our experiments, both the NSP and our model were used to predict outcomes on these datasets. In the third dataset, where token length exceeds the NSP model's limit, we truncated the conversation for NSP input, while our model used the entire conversation. Table 1 shows the results. NSP performs similarly to our model for small token gaps, but as the gap widens, our model outperforms NSP significantly. With token gaps surpassing 512, NSP's results become unreliable due to excluding the target sentence from its input. In contrast, our model maintains high accuracy. This experiment underscores our model's superior performance in managing conversations of varying lengths, achieving state-of-the-art results.


\section{Conclusion}
With the rapid development of large language models (LLMs), the effective utilization of LLMs in various business scenarios has become an important issue. In this paper, we propose a method that ensures user conversations with LLMs remain focused on fixed topics. This method is based on the introduction of non-linear transformations and attention mechanisms through an extension of Naive Bayes. Experimental results across various scenarios consistently demonstrate that our approach outperforms traditional methods. We believe this method will be highly beneficial for using LLMs in topic-constrained scenarios.

\bibliography{citation}

@article{llm-survey,
  title={A Survey of Large Language Models},
  author={Wayne Xin Zhao and Kun Zhou and Junyi Li and Tianyi Tang and Xiaolei Wang and Yupeng Hou and Yingqian Min and Beichen Zhang and Junjie Zhang and Zican Dong and Yifan Du and Chen Yang and Yushuo Chen and Zhipeng Chen and Jinhao Jiang and Ruiyang Ren and Yifan Li and Xinyu Tang and Zikang Liu and Peiyu Liu and Jian-Yun Nie and Ji-Rong Wen},
  journal={arXiv},
  volume={},
  number={2303.18223},
  pages={},
  year={2023},
  publisher={}
}

@article{llm-evaluation,
  title={A Survey on Evaluation of Large Language Models},
  author={Yupeng Chang and Xu Wang and Jindong Wang and Yuan Wu and Linyi Yang and Kaijie Zhu and Hao Chen and Xiaoyuan Yi and Cunxiang Wang and Yidong Wang and Wei Ye and Yue Zhang and Yi Chang and Philip S. Yu and Qiang Yang and Xing Xie},
  journal={ACM Transactions on Intelligent Systems and Technology},
  volume={},
  number={},
  pages={DOI:10.1145/3641289},
  year={2024},
  publisher={}
}

@article{powerpoint-llm,
  title={PPTC Benchmark: Evaluating Large Language Models for PowerPoint Task Completion.},
  author={Yiduo Guo and Zekai Zhang and Yaobo Liang and Dongyan Zhao and Nan Duan},
  journal={arXiv},
  volume={},
  number={2311.01767},
  pages={},
  year={2023},
  publisher={}
}

@article{llm-code1,
  title={Expectation vs. Experience: Evaluating the Usability of Code Generation Tools Powered by Large Language Models},
  author={P Vaithilingam and T Zhang and EL Glassman },
  journal={CHI EA '22: Extended Abstracts of the 2022 CHI Conference on Human Factors in Computing Systems},
  volume={332},
  number={},
  pages={1-7},
  year={2022},
  publisher={}
}

@article{llm-code2,
  title={Planning with Large Language Models for Code Generation},
  author={Shun Zhang and Zhenfang Chen and Yikang Shen and Mingyu Ding and Joshua B. Tenenbaum and Chuang Gan },
  journal={arXiv},
  volume={},
  number={2303.05510},
  pages={},
  year={2023},
  publisher={}
}

@article{llm-data2text,
  title={A Survey on Neural Data-to-Text Generation},
  author={Yupian Lin and Tong Ruan and Jingping Liu and Haofen Wang},
  journal={IEEE Transactions on Knowledge and Data Engineering},
  volume={},
  number={},
  pages={1-20},
  year={2023},
  publisher={}
}

@article{llm-hallucination,
  title={Survey of Hallucination in Natural Language Generation},
  author={Ziwei Ji and Nayeon Lee and Rita Frieske and Tiezheng Yu and Dan Su and Yan Xu and Etsuko Ishii and Ye Jin Bang and Andrea Madotto and Pascale Fung},
  journal={ACM Computing Surveys},
  volume={55},
  number={248},
  pages={1-38},
  year={2023},
  publisher={}
}

@article{llm-offensive,
  title={Towards Understanding and Mitigating Social Biases in Language Models},
  author={Paul Pu Liang and Chiyu Wu and Louis-Philippe Morency and Ruslan Salakhutdinov},
  journal={Proceedings of the 38th International Conference on Machine Learning, PMLR},
  volume={139},
  number={},
  pages={6565-6576},
  year={2021},
  publisher={}
}

@article{prompt-injection,
  title={Not what you've signed up for: Compromising Real-World LLM-Integrated Applications with Indirect Prompt Injection},
  author={Kai Greshake and Sahar Abdelnabi and Shailesh Mishra and Christoph Endres and Thorsten Holz and Mario Fritz},
  journal={arXiv},
  volume={},
  number={2302.12173},
  pages={},
  year={2023},
  publisher={}
}

@article{adversarial,
  title={Survey of Vulnerabilities in Large Language Models Revealed by Adversarial Attacks},
  author={Erfan Shayegani and Md Abdullah Al Mamun and Yu Fu and Pedram Zaree and Yue Dong and Nael Abu-Ghazaleh},
  journal={arXiv},
  volume={},
  number={2310.10844},
  pages={},
  year={2023},
  publisher={}
}

@article{customer-service1,
  title={Universal model in online customer service},
  author={Shu-Ting Pi and Cheng-Ping Hsieh and Qun Liu and Yuying Zhu},
  journal={Companion Proceedings of the ACM Web Conference 2023},
  volume={},
  number={},
  pages={878–885},
  year={2023},
  publisher={}
}

@article{customer-service2,
  title={Contact complexity in customer service},
  author={ Shu-Ting Pi and Michael Yang and Qun Liu},
  journal={arXiv},
  volume={},
  number={2402.15655},
  pages={},
  year={2024},
  publisher={}
}

@article{customer-service3,
  title={Teacher-Student Learning on Complexity in Intelligent Routing},
  author={Shu-Ting Pi and Michael Yang and Yuying Zhu and Qun Liu},
  journal={arXiv},
  volume={},
  number={2402.15665},
  pages={},
  year={2024},
  publisher={}
}

@article{customer-service4,
  title={Uncovering Customer Issues through Topological Natural Language Analysis },
  author={ Shu-Ting Pi and Sidarth Srinivasan and Yuying Zhu and Michael Yang and Qun Liu},
  journal={arXiv},
  volume={},
  number={2403.00804},
  pages={},
  year={2024},
  publisher={}
}

@article{NLU,
  title={Natural Language Processing Advancements By Deep Learning: A Survey},
  author={ Amirsina Torfi and Rouzbeh A. Shirvani and Yaser Keneshloo and Nader Tavaf and Edward A. Fox},
  journal={arXiv},
  volume={},
  number={2003:01200},
  pages={},
  year={2003},
  publisher={2020}
}

@article{transformer,
  title={Attention Is All You Need},
  author={Ashish Vaswani and Noam Shazeer and Niki Parmar and Jakob Uszkoreit and Llion Jones and Aidan N. Gomez and Lukasz Kaiser and Illia Polosukhin},
  journal={Advances in neural information processing systems},
  volume={30},
  number={},
  pages={},
  year={2017},
  publisher={}
}

@article{bert,
  title={BERT: Pre-training of Deep Bidirectional Transformers for Language Understanding},
  author={Jacob Devlin and Ming-Wei Chang and Kenton Lee, Kristina Toutanova},
  journal={Advances in neural information processing systems},
  volume={30},
  number={},
  pages={},
  year={2017},
  publisher={}
}

@article{naive_bayes,
  title={ An empirical study of the naive Bayes classifier},
  author={Irina Rish},
  journal={IJCAI 2001 workshop on empirical methods in artificial intelligence},
  volume={3},
  number={22},
  pages={41-46},
  year={2001},
  publisher={}
}

@article{nsp1,
  title={ Next sentence prediction helps implicit discourse relation classification within and across domains},
  author={Wei Shi and Vera Demberg},
  journal={Proceedings of the 2019 Conference on Empirical Methods in Natural Language Processing and the 9th International Joint Conference on Natural Language Processing (EMNLP-IJCNLP)},
  volume={},
  number={},
  pages={5790-5796},
  year={2019},
  publisher={}
}

@article{nsp2,
  title={ NSP-BERT: A Prompt-based Few-Shot Learner Through an Original Pre-training Task--Next Sentence Prediction},
  author={Yi Sun and Yu Zheng and Chao Hao and Hangping Qiu},
  journal={arXiv},
  volume={},
  number={arXiv:2109.03564},
  pages={},
  year={2021},
  publisher={}
}

@article{isolation_forest1,
  title={Isolation forest},
  author={Fei Tony Liu and Kai Ming Ting and Zhi-Hua Zhou},
  journal={IEEE International Conference on Data Mining},
  volume={},
  number={},
  pages={10472172},
  year={2008},
  publisher={}
}

@article{isolation_forest2,
  title={Isolation forest based anomaly detection},
  author={Fei Tony Liu and Kai Ming Ting and Zhi-Hua Zhou},
  journal={ACM Transactions on Knowledge Discovery from Data},
  volume={6},
  number={1},
  pages={1-39},
  year={2012},
  publisher={}
}

@article{conversation_bert,
  title={},
  author={DeepPavlov.ai},
  journal={https://huggingface.co/DeepPavlov/bert-base-cased-conversational},
  volume={},
  number={},
  pages={},
  year={2021},
  publisher={}
}

@article{sentence-bert,
  title={SentenceBERT: Sentence embeddings using siamese BERTnetworks .},
  author={Nils Reimers and Iryna Gurevych},
  journal={Proceedings of the 2019 Conference on Empirical Methods in Natural Language Processing and the 9th International Joint Conference on Natural Language Processing},
  volume={},
  number={},
  pages={3982–3992 (Also see, https://www.sbert.net/)},
  year={2019},
  publisher={}
}

@article{albert,
  title={ALBERT: A Lite BERT for Self-supervised Learning of Language Representations},
  author={Zhenzhong Lan and Mingda Chen and Sebastian Goodman and Kevin Gimpel and Piyush Sharma and Radu Soricut},
  journal={arXiv},
  volume={},
  number={1909.11942},
  pages={},
  year={2019},
  publisher={}
}

@article{erine,
  title={ERNIE: Enhanced Language Representation with Informative Entities},
  author={Zhengyan Zhang and Xu Han and Zhiyuan Liu and Xin Jiang and Maosong Sun and Qun Liu},
  journal={arXiv},
  volume={},
  number={1905.07129},
  pages={},
  year={2019},
  publisher={}
}

@article{erine2,
  title={ERNIE 2.0: A Continual Pre-Training Framework for Language Understanding},
  author={Yu Sun and Shuohuan Wang and Yukun Li and Shikun Feng and Hao Tian and Hua Wu and Haifeng Wang},
  journal={Proceedings of the AAAI Conference on Artificial Intelligence},
  volume={34},
  number={05},
  pages={8968-8975},
  year={2020},
  publisher={}
}

\end{document}